\documentclass{article}
\usepackage{ijcai20}

\usepackage{times}
\usepackage{soul}
\usepackage{url}
\usepackage[hidelinks]{hyperref}
\usepackage[utf8]{inputenc}
\usepackage[small]{caption}
\usepackage{graphicx}
\usepackage{amsmath}
\usepackage{amsthm}
\usepackage{booktabs}
\usepackage{algorithm}
\usepackage{algorithmic}
\usepackage{subcaption}

\title{Extending Class Activation Mapping Using Gaussian Receptive Field}

\author{
Bum Jun Kim$^1$
\and
Gyogwon Koo$^1$\and
Hyeyeon Choi$^1$\And
Sang Woo Kim$^{1}$\footnote{Contact Author}
\affiliations
$^1$Department of Electrical Engineering, Pohang University of Science and Technology, Pohang, 37673, South Korea\\
\emails
\{kmbmjn, ggkoo99, hyeyeon, swkim\}@postech.edu
}
\begin{document}

\maketitle

\begin{abstract}
 This paper addresses the visualization task of deep learning models. To improve Class Activation Mapping (CAM) based visualization method, we offer two options. First, we propose Gaussian upsampling, an improved upsampling method that can reflect the characteristics of deep learning models. Second, we identify and modify unnatural terms in the mathematical derivation of the existing CAM studies. Based on two options, we propose Extended-CAM, an advanced CAM-based visualization method, which exhibits improved theoretical properties. Experimental results show that Extended-CAM provides more accurate visualization than the existing methods. 
\end{abstract}

\section{Introduction} \label{1 Introduction}

Deep Convolutional Neural Networks (DCNNs) have exhibited remarkable accuracy in various image processing fields such as object detection \cite{redmon2016you,girshick2015fast}, semantic segmentation \cite{long2015fully,chen2017deeplab}, and monocular depth estimation \cite{godard2017unsupervised,fu2018deep}. DCNNs have been confirmed in various studies to possess strong modeling abilities in general machine learning problems. DCNN, however, is a black-box model and does not provide the cause or process for the results it outputs. As practical applications require stability and understanding, researches have been conducted to visualize the inner behavior of DCNN.

The visualization methods of DCNN are largely divided into two popular branches. The first is the gradient-based methods \cite{zeiler2014visualizing,springenberg2014striving}. They produce activation by flowing the gradient of DCNN to the learned weights. The gradient-based methods mainly pay attention to detail. The second is the Class Activation Mapping (CAM) methods \cite{zhou2016learning,selvaraju2017grad,chattopadhay2018grad}. CAM methods operate on feature maps and represent rough shapes. Each algorithm is developed independently, but recently, \cite{selvaraju2017grad} mixed the two methods to complement their advantages and disadvantages.

In this paper, several problems in existing CAM studies are identified. We propose Extended-CAM, a new visualization method that improves on these problems. The main contributions of this paper are summarized as follows:

\begin{enumerate}
	\item We propose an improved upsampling method, Gaussian upsampling. The Gaussian upsampling offers a natural way for upsampling CAM and reflects the characteristics of DCNN. We argue that bilinear upsampling is an inappropriate upsampling method for CAM. \\
	
	\item Problematic mathematical derivation on existing CAM studies are identified. We discuss the correct mathematical derivation for DCNN that exhibits non-linearities. \\
	
	\item Combining the two options, we propose Extended-CAM, a new CAM-based visualization method. We verified that our Extended-CAM provides more accurate visualization than the existing methods. \\
\end{enumerate}

This paper is organized as follows. Section \ref{2 Related Work} reviews the literature on CAM. Section \ref{3 Method} describes the two components that consist of Extended-CAM, Gaussian upsampling and modified mathematical derivation. We verify the accuracy of the visualization mask with experiments in Section \ref{4 Experiments} and discuss underlying meaning in Section \ref{5 Discussion}.

\begin{figure}[t!]
	\centering
	\begin{subfigure}{.23\textwidth}
		\centering
		% include first image
		\includegraphics[width=\linewidth]{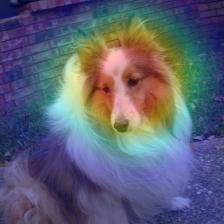} 
	\end{subfigure}
	\begin{subfigure}{.23\textwidth}
		\centering
		% include second image
		\includegraphics[width=\linewidth]{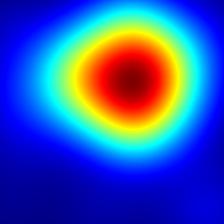} 
	\end{subfigure}
	\caption{An example of visualizing DCNN's inner behavior for a given image using our Extended-CAM.} 	\label{fig:dog}
\end{figure}

\section{Related Work} \label{2 Related Work}

This section reviews existing CAM studies. While the CAM approach is general for DCNNs, we mainly focus on DCNN for classification problems of 2D images such as the VGG models. DCNN consists of a feature extractor and classifier. The feature extractor in front of DCNN consists of convolutional layers and pooling layers, obtaining feature maps from the input image. The final output of the feature extractor is the last feature map $A_{ijk}$, where $i, j$ are spatial indices and $k$ is channel index. The last feature map $A_{ijk}$ passes through the Fully-Connected layers, which forms classifier at the back, to obtain a score $y^c$. Finally, $y^c$ is passed through the softmax layer to obtain a probability for classification.

The goal of the CAM methods is to obtain $L_{ij}^c$ which satisfies $y^c=\sum_{i,j}L_{ij}^c$ . Here, $i, j$ are the indices of $A_{ijk}$, the last feature map of DCNN. In other words, the CAM aims to calculate the contribution to prediction $y^c$ in units of $i, j$. 

Original CAM \cite{zhou2016learning} obtains $L_{ij}^c$ by replacing the classifier architecture of DCNN after feature extractor with linear layers. The linear layers consist of Global Average Pooling (GAP) and Fully-Connected layer (FC). At this time, the last feature map $A_{ijk}$ and prediction $y^c$ are represented in a linear relationship. Specifically, GAP outputs, $F_k=\sum_{i,j}A_{ijk}$, and FC that holds weight $w_k^c$ outputs $y^c=\sum_k{w_k^c F_k}$, so we derive,

\begin{align}
y^c = \sum_k{w_k^c F_k} = \sum_k{w_k^c \sum_{i,j}} A_{ijk}=\sum_{i,j}\sum_k{w_k^c A_{ijk}}.
\end{align}%

As the goal is to find $L_{ij}^c$ that satisfies $y^c=\sum_{i,j}L_{ij}^c$, we obtain $L_{ij}^c=\sum_k{w_k^c A_{ijk}}$.

The disadvantages of original CAM lie in the linear layers. If linear layers replace the classifier architecture, retraining of DCNN is required and non-linearity of classifier vanishes. To enable the direct use of the CAM approach in DCNN without any modification of the architecture, Grad-CAM/Grad-CAM++ \cite{selvaraju2017grad,chattopadhay2018grad} have been proposed. As in the original CAM, they also frame the form $L_{ij}^c=\sum_k{w_k^c A_{ijk}}$ but use mathematical derivation to obtain the coefficient $w_k^c$ under the assumption of the non-linear classifier. The resulting Grad-CAM equation is:

 \begin{align}
 w_k^c=\frac{1}{Z}\sum_{i,j}{\frac{\partial y^c}{\partial A_{ijk}}}.
 \end{align}%
 
\noindent where $Z$ is the area of the last feature map. Grad-CAM++ adds a term $\alpha_{ijk}^c$ to generalize Grad-CAM and uses,

\begin{align}
w_k^c &=\sum_{i,j}\alpha_{ijk}^c ReLU({\frac{\partial y^c}{\partial A_{ijk}}}),
\end{align}%

\noindent where $\alpha_{ijk}^c = \frac{\frac{\partial^2 y^c}{\partial A_{ijk}^2}}{2\frac{\partial^2 y^c}{\partial A_{ijk}^2}+\sum_{a, b}A_{abk}{\frac{\partial^3 y^c}{\partial A_{ijk}^3}}}$. By combining $w_k^c$ and $A_{ijk}$, we obtain, $L_{ij}^c=\sum_k{w_k^c A_{ijk}}$. 

\section{Method} \label{3 Method}

\subsection{Gaussian Upsampling}
In DCNN, downsampled feature maps appear due to the pooling layers or the strided convolution layers. Accordingly, the units of last feature map $i, j$ are different from pixel-level units. Assume a $224 \times 224$ image $I$ is given to a DCNN. Using a CAM method, we obtain a grid-level CAM like $14 \times 14$ $L_{ij}^c$. To obtain pixel-level CAM $224 \times 224$ $L_{xy}^c$, where ${x, y}$ are units of pixel-level, an upsampling method is required.

Here, bilinear upsampling method is used in CAM, Grad-CAM, and Grad-CAM++. Nevertheless, bilinear upsampling is a simple method, which linearly interpolates only through adjacent values. The fatal problem of bilinear upsampling is that it fails to link the physical meaning between the grid-level CAM $14 \times 14$ $L_{ij}^c$ and the pixel-level CAM $224 \times 224$ $L_{xy}^c$. We want to revise the upsampling method to reflect how the grid and pixel link, i.e., how many pixels a grid-level cell covers.

Such a link can be investigated from the nature of the convolutional operation. Indeed, a cell value of the feature map is determined only by a local region of the pixel-level, which is known as a receptive field. Meanwhile, \cite{luo2016understanding} found effective receptive field, which is different from the previously known square type receptive field. By checking the pixel that is actually activated, the shape of the effective receptive field is shown to be a 2D Gaussian.

 \begin{figure}[t] %%% t: top, b: bottom, h: here
	\begin{center}
		\includegraphics[width=1.0\linewidth]{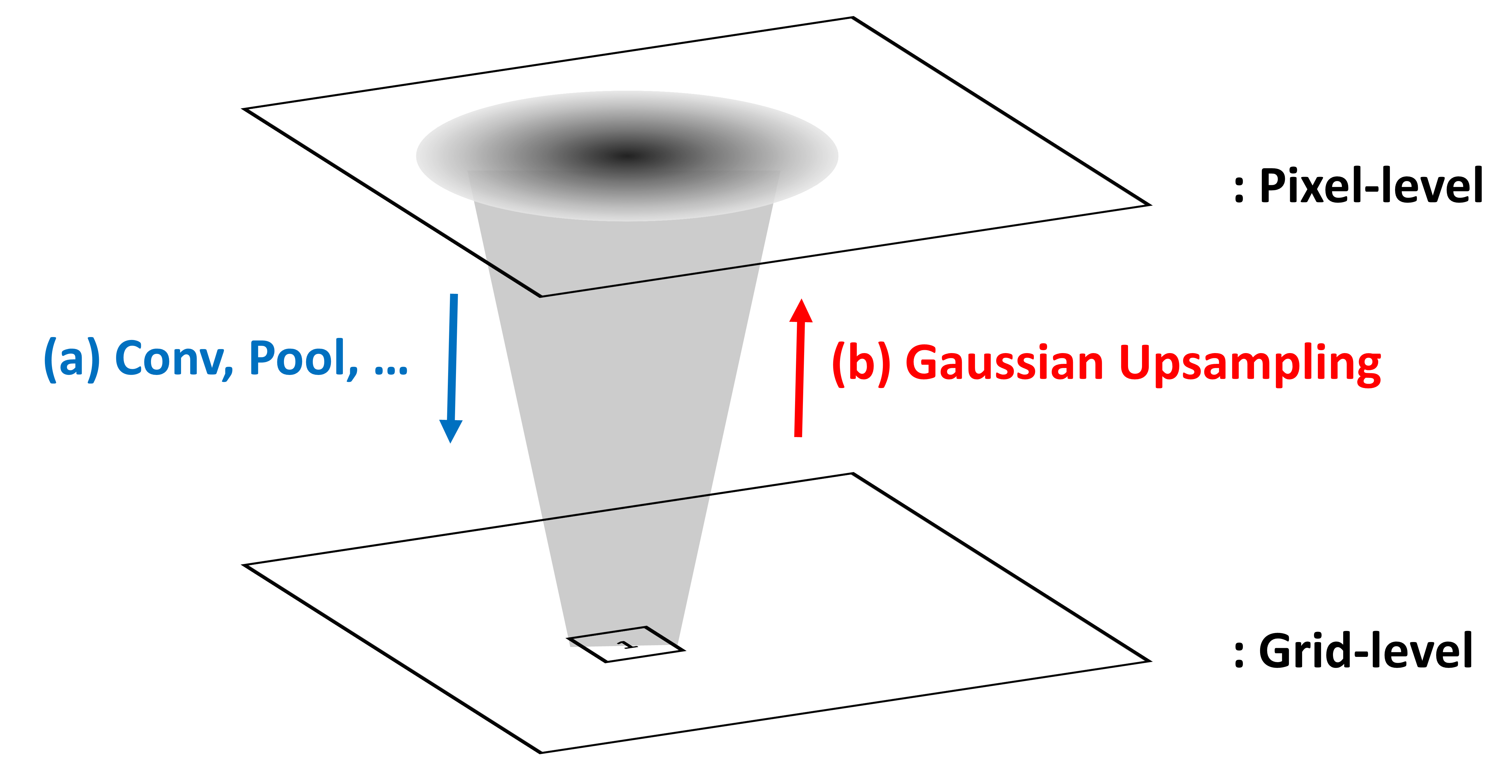}
	\end{center}
	\caption{(a) When the pixel-level effective receptive field region passes through convolution and pooling layers, the value of one grid-level cell is determined. (b) Conversely, to interpret the grid-level value in the pixel-level, the value of one cell in the grid must spray a 2D Gaussian of the effective receptive field on the pixel-level.}
	\label{fig:erfpixelgrid}
\end{figure}

Thus, a pixel-level region of effective receptive field determines the value of a grid-level cell as it passes through a feature extractor of the DCNN. Conversely, the pixel-level region covered by a grid-level cell would appear as the 2D Gaussian of the effective receptive field (Fig. \ref{fig:erfpixelgrid}). We describe this as spraying 2D Gaussian on the pixel-level. Therefore, to transform a grid-level CAM to hold meaning at the pixel-level, all cell values in the $14 \times 14$ grid must spray 2D Gaussians on the pixel-level.

Now, we investigate the CAM value of a pixel. The CAM value of a pixel is determined by the combination of Gaussian sprays of all the cells in the grid-level CAM. Due to the characteristics of Gaussian spray, a pixel is sprayed much from the close cell, and less from the far cell. We express this as a combination of 2D Gaussian, as follows.

\begin{align}
L_{xy}^c = \sum_{i,j}{L_{ij}^c exp\Bigg[-\Bigg\{\frac{(x-\frac{w}{u}i)^2}{2\hat\sigma_X^2}
	+\frac{(y-\frac{h}{v}j)^2}{2\hat\sigma_Y^2}\Bigg\}\Bigg]}, \label{eqn:gaussianup}
\end{align}%

\noindent where $u, v = 14$, $w, h = 224$, and $\hat\sigma_X$, and $\hat\sigma_Y$ are the standard deviations for 2D Gaussian of the effective receptive field. We call this improved upsampling method that links physical meanings of grid and pixel, {\it Gaussian upsampling}.

Although we use the term “upsampling”, Gaussian upsampling differs from the conventional upsampling method. Gaussian upsampling results in $16 \times$ upsampling, but it aims to naturally link the physical meaning of the pixel and grid. Gaussian upsampling does not aim to increase resolution by making interpolation more accurate. According to Gaussian upsampling, the CAM value of a pixel is determined by a Gaussian combination of the entire grid. Even if a grid cell is far away, a pixel may be influenced. Thus, conventional upsampling methods that interpolate only through adjacent values are completely different from Gaussian upsampling. Conversely, such a conventional upsampling method is unsuitable for the upsampling of CAM because distant, non-adjacent grid cell values cannot influence the corresponding pixel.

\subsection{Modified Mathematical Derivation}
CAM replaced DCNN's classifier architecture with linear layers of Global Average Pooling (GAP) and a Fully-Connected (FC) layer. However, the neural network is essentially a non-linear model. To apply the CAM approach and allow non-linearity of DCNN, Grad-CAM/Grad-CAM++ have been proposed.

However, Grad-CAM/Grad-CAM++ hold two common problems. First, even Grad-CAM/Grad-CAM++ assume linear layers of GAP and FC in mathematical derivation. This is unsuitable for assuming non-linear layers. Second, Grad-CAM/Grad-CAM++ use $w_k^c$ as the coefficient of $A_{ijk}$, and it is assumed to be independent of $i, j$. We will demonstrate that this is also an unnecessary assumption and the use of $w_{ijk}^c$, which include dependent information on $i, j$, is more natural.

Following mathematical derivation we provide improves on these two problems. In DCNN, the last feature map $A_{ijk}$ passes through Fully-Connected layers, which outputs the prediction score $y^c$. We represent the Fully-Connected layers by a non-linear function $f$.

\begin{align}
y^c &= f(A_{111}, A_{112}, ...) \label{eqn:ourderiv1} \\
&= C + (\frac{\partial y^c}{\partial A_{111}}A_{111} + ...) 
+ (\frac{\partial^2 y^c}{\partial A_{111}^2}A_{111}^2 + ...)
+ O(A^3).\label{eqn:ourderiv2}
\end{align}%

The function $f$ is non-linear because the activation function of DCNN holds non-linearity. We note the form of activation functions such as ReLU, LReLU, sigmoid, and tanh \cite{nair2010rectified,maas2013rectifier}. We assume that piecewise-linear function can approximate them. Using this approximation, all terms of quadratic and higher degree in Equation \ref{eqn:ourderiv2} are eliminated. We confirmed that the constant term $C$ was insignificant and ignored it. So, we approximate the non-linear function $f$ only through the first order.

\begin{align}
	y^c = \sum_{i,j,k}{\frac{\partial y^c}{\partial A_{ijk}}A_{ijk}}.
\end{align}%

The goal of the CAM approach is to obtain $L_{ij}^c$ such that $y^c=\sum_{i,j}L_{ij}^c$. So,

\begin{align}
	L_{ij}^c=\sum_k\frac{\partial y^c}{\partial A_{ijk}} A_{ijk}. \label{eqn:wijkc}
\end{align}%

In the context of existing CAM studies which use $w_{k}^c$, Equation \ref{eqn:wijkc} can be interpreted as using the coefficient $w_{ijk}^c=\frac{\partial y^c}{\partial A_{ijk}}$, which includes $i, j$ units. We can understand that the use of $w_{ijk}^c$, which consider $i, j$ units, is mathematically natural. It is unnatural to additionally assume that the coefficient is independent of $i, j$ and to use $w_{k}^c$, which is averaged over $i, j$, like Grad-CAM. The meaning of $w_{ijk}^c$ will be discussed again in the discussion section.

On the other hand, additional post-processing is applied in the existing CAM studies. Grad-CAM applies $ReLU$ on $L_{ij}^c$ and clips the negative value to 0 to ignore the negative contribution. Grad-CAM++ also takes $ReLU$ on the gradient term to ignore the negative gradient. These manipulations using $ReLU$ lead to several problems: the information of negative value is lost, an unnatural formula is created, and the design choice of where to assign $ReLU$ is required. However, our method uses Gaussian upsampling, so even if a negative value appears in $L_{ij}^c$, it is smoothed by blending with the surrounding value. Thus, the information of negative $L_{ij}^c$ can be naturally reflected in the surrounding value. Experimenting with various design choices that apply $ReLU$ on $L_{ij}^c$ and the gradient term, we confirmed that the original Equation \ref{eqn:wijkc} without any $ReLU$ yielded the best results.

\section{Experiments} \label{4 Experiments}
\subsection{Estimation of Effective Receptive Field}
\begin{figure}[h] %%% t: top, b: bottom, h: here
	\begin{center}
		\includegraphics[width=0.6\linewidth]{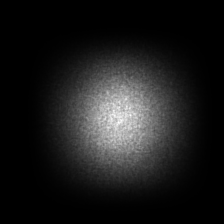}
	\end{center}
	\caption{The effective receptive field of the VGG-16 model. The bright areas are highly activated areas due to large gradients, and the dark areas are areas that are not.}
	\label{fig:erfvgg}
\end{figure}

To use Gaussian upsampling, we need to estimate $\hat\sigma_X$, $\hat\sigma_Y$, the standard deviations of 2D Gaussian of the effective receptive field. First, we obtained the effective receptive field of the last feature map of the VGG-16 model \cite{simonyan2014very}. This can be achieved by obtaining the gradient of an image for the last feature map and averaging it over the various images (Fig. \ref{fig:erfvgg}).

\begin{figure*}[ht!]
	\centering
	\begin{subfigure}{.33\textwidth}
		\caption{Grad-CAM}
		\centering
		% include first image
		\includegraphics[width=.47\linewidth]{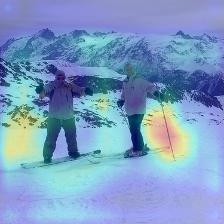} 
		\includegraphics[width=.47\linewidth]{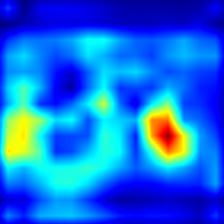} 
	\end{subfigure}
	\begin{subfigure}{.33\textwidth}
		\caption{Grad-CAM++}
		\centering
		% include first image
		\includegraphics[width=.47\linewidth]{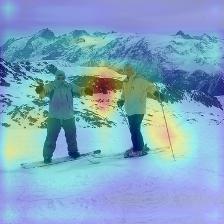} 
		\includegraphics[width=.47\linewidth]{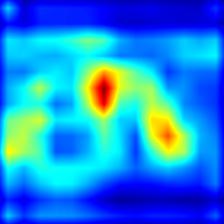} 
	\end{subfigure}
	\begin{subfigure}{.33\textwidth}
		\caption{Extended-CAM}
		\centering
		% include first image
		\includegraphics[width=.47\linewidth]{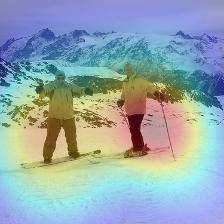} 
		\includegraphics[width=.47\linewidth]{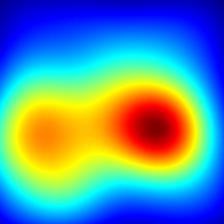} 
	\end{subfigure}
	\hfill
	
	\begin{subfigure}{.33\textwidth}
		\centering
		% include first image
		\includegraphics[width=.47\linewidth]{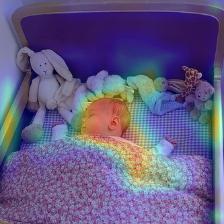} 
		\includegraphics[width=.47\linewidth]{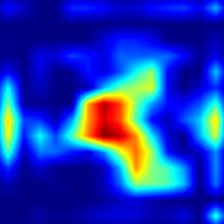} 
	\end{subfigure}
	\begin{subfigure}{.33\textwidth}
		\centering
		% include first image
		\includegraphics[width=.47\linewidth]{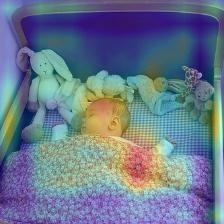} 
		\includegraphics[width=.47\linewidth]{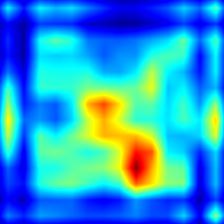} 
	\end{subfigure}
	\begin{subfigure}{.33\textwidth}
		\centering
		% include first image
		\includegraphics[width=.47\linewidth]{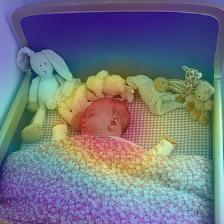} 
		\includegraphics[width=.47\linewidth]{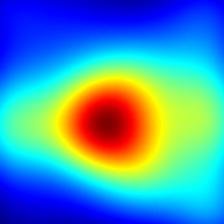} 
	\end{subfigure}
	\hfill
	
	\begin{subfigure}{.33\textwidth}
		\centering
		% include first image
		\includegraphics[width=.47\linewidth]{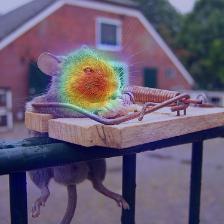} 
		\includegraphics[width=.47\linewidth]{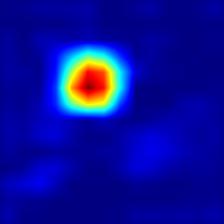} 
	\end{subfigure}
	\begin{subfigure}{.33\textwidth}
		\centering
		% include first image
		\includegraphics[width=.47\linewidth]{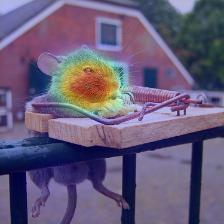} 
		\includegraphics[width=.47\linewidth]{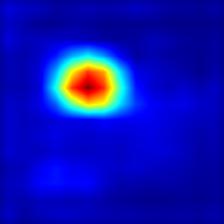} 
	\end{subfigure}
	\begin{subfigure}{.33\textwidth}
		\centering
		% include first image
		\includegraphics[width=.47\linewidth]{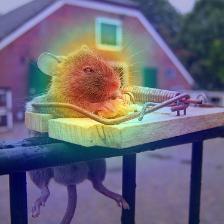} 
		\includegraphics[width=.47\linewidth]{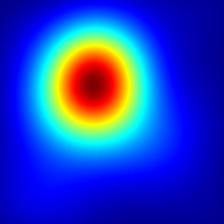} 
	\end{subfigure}
	
	\caption{Visualization examples of existing CAM methods and Extended-CAM. We confirm that the visualization results of Extended-CAM represent more important areas.}
	\label{fig:visexamples}
\end{figure*}

 We used the {\tt LmFit} library \cite{newville2016lmfit} to fit the effective receptive field into 2D Gaussian. The unused pixels which appear as microscopic holes in the effective receptive field have negative values. We observed that ignoring them does not affect the fitting results.

\begin{table}[h]
	\centering
	\begin{tabular}{lr}  
		\toprule
		Item  & Value  \\
		\midrule
		$R^2$       & 0.9905   \\
		$\hat\sigma_X$    & 31.7797  \\
		$\hat\sigma_Y$    & 33.3606  \\
		\bottomrule
	\end{tabular}
    \caption{2D Gaussian fitting result of the effective receptive field. The $ R ^ 2 $ value which is close to 1 indicates that the effective receptive field well matches to the 2D Gaussian shape. The resulting $\hat\sigma_X$, $\hat\sigma_Y$ are used for Gaussian upsampling. }
	\label{tab:erffitting}
\end{table}

\begin{table*}[htb!]
	\centering
	\begin{tabular}{lrrr}  
		\toprule
		Method & Grad-CAM & Grad-CAM++ & Extended-CAM (Ours) \\
		\midrule
		Average Drop \% (Lower Better)& 46.56 & 36.84 & {\bf 21.89} \\
		\% Increase (Higher Better) & 13.42 & 17.05 & {\bf 28.72} \\
		\bottomrule
	\end{tabular}
    \caption{Comparison of visualization performance of CAM-based methods.}
	\label{tab:compareresults}
\end{table*}

\begin{table*}[htb!]
	\centering
	\begin{tabular}{lrrr}  
		\toprule
		Method & Grad-CAM & Grad-CAM++ & Extended-CAM (Ours) \\
		\midrule
		Average Drop \% (Lower Better)& 47.94 & 45.60 & {\bf 39.86} \\
		\% Increase (Higher Better) & 16.39 & 15.10 & {\bf 20.72} \\
		\bottomrule
	\end{tabular}
    \caption{Comparison of visualization performance of CAM-based methods with relative masking.}
	\label{tab:compareresultsrelative}
\end{table*}

The fitting results showed that the $R^2$ value is quite close to 1, indicating that the effective receptive field matches well to the shape of the 2D Gaussian (Table \ref{tab:erffitting}). The resulting $\hat\sigma_X$, $\hat\sigma_Y$ are used in the Gaussian upsampling.

\subsection{Visualization with Extended-CAM}
We implemented Extended-CAM, which uses Gaussian upsampling and the modified mathematical equation using $w_{ijk}^c$. We obtained several examples of visualization from Extended-CAM and existing CAM methods (Fig. \ref{fig:visexamples}). This visualization represents important parts for the VGG-16 model to classify a given image. The visualization results of Grad-CAM/Grad-CAM++ often failed to accurately represent important areas. Also, they suffer from various artifacts such as cross or rhombus patterns in the visualization results. These artifacts appear when bilinear upsampling is applied in 2D. In contrast, the visualization of Extended-CAM appears more accurate and more natural without any artifacts. Besides, Extended-CAM represents less noise in other areas than Grad-CAM/Grad-CAM++. 

Following the same experiments in \cite{chattopadhay2018grad}, we systematically evaluated the accuracy of these visualization masks. They measured the confidence change of DCNN when an image is masked. These DCNN-based evaluation methods are often used in a variety of fields because DCNN's characteristics resemble human perception \cite{johnson2016perceptual,zhang2018unreasonable,salimans2016improved}. First, we masked the original image $I$ with $L_{xy}^c$ to obtain the masked image $E^c$. The confidences that DCNN outputs for $I$ and $E^c$ are then compared. If $L_{xy}^c$ is an inaccurate visualization, it masks out important parts in the image, resulting in the dropped confidence. Conversely, the less the confidence drops, the better the visualization is. The tendency to increase confidence also represents an accurate visualization as it removes unnecessary parts in the image. To compare the confidences, we used two indicators, Average Drop \% and \% Increase. We measured them on average using the PASCAL VOC 2007 validation dataset \cite{everingham2007pascal}.

On both Average Drop \% and \% Increase, Extended-CAM outperforms the existing CAM methods (Table \ref{tab:compareresults}). The result indicates that confidence drops less and the tendency to increase is larger. These confidence changes mean that important parts of the image remain well when applying the visualization mask suggested by Extended-CAM. In other words, the visualization of Extended-CAM highlights more important parts of the image than the existing CAM methods.

\begin{figure*}[htb!]
	\begin{subfigure}[t]{.5\textwidth}
		\centering
		\includegraphics[width=.9\linewidth]{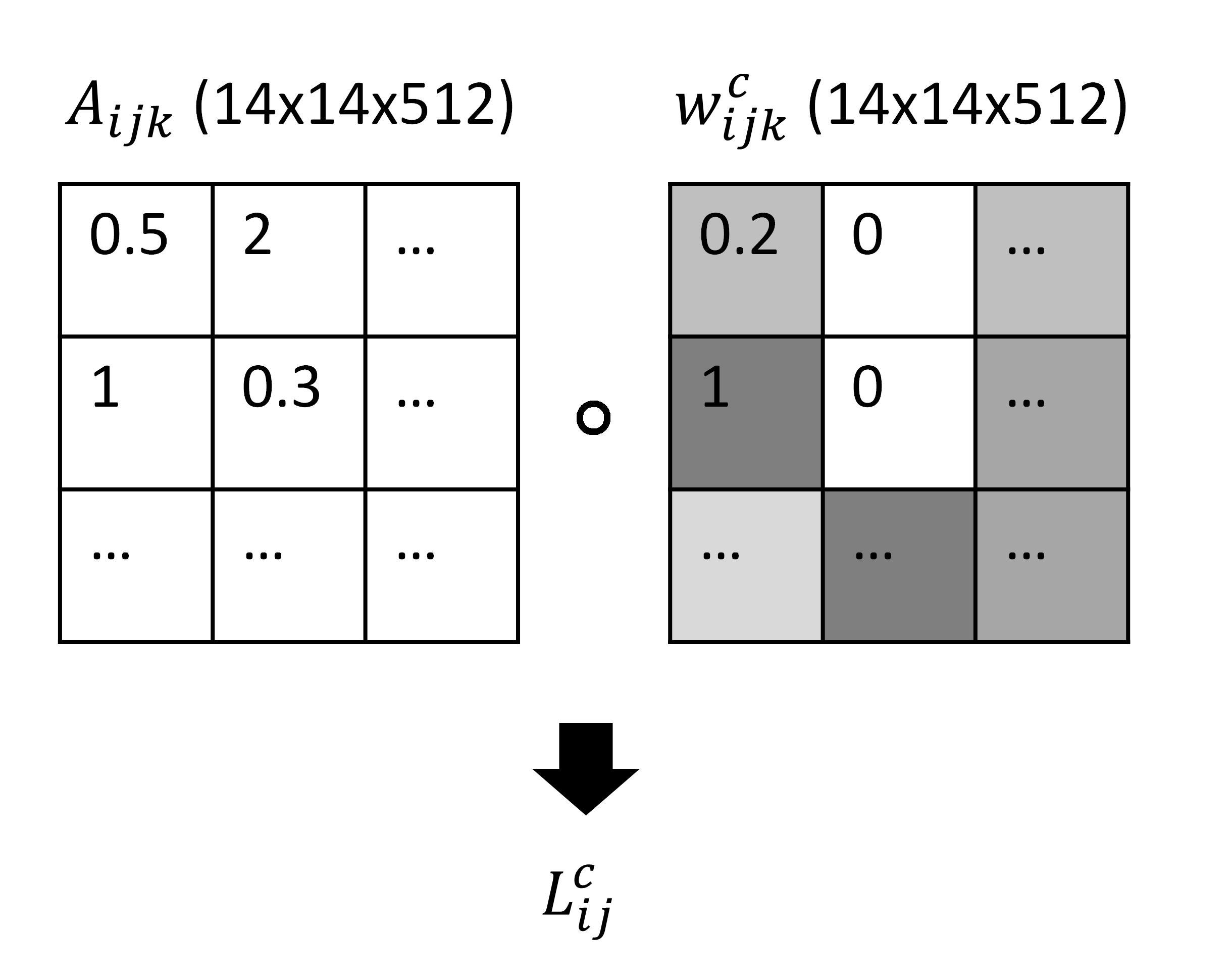} 
	\end{subfigure}
	\begin{subfigure}[t]{.5\textwidth}
		\centering
		\includegraphics[width=.9\linewidth]{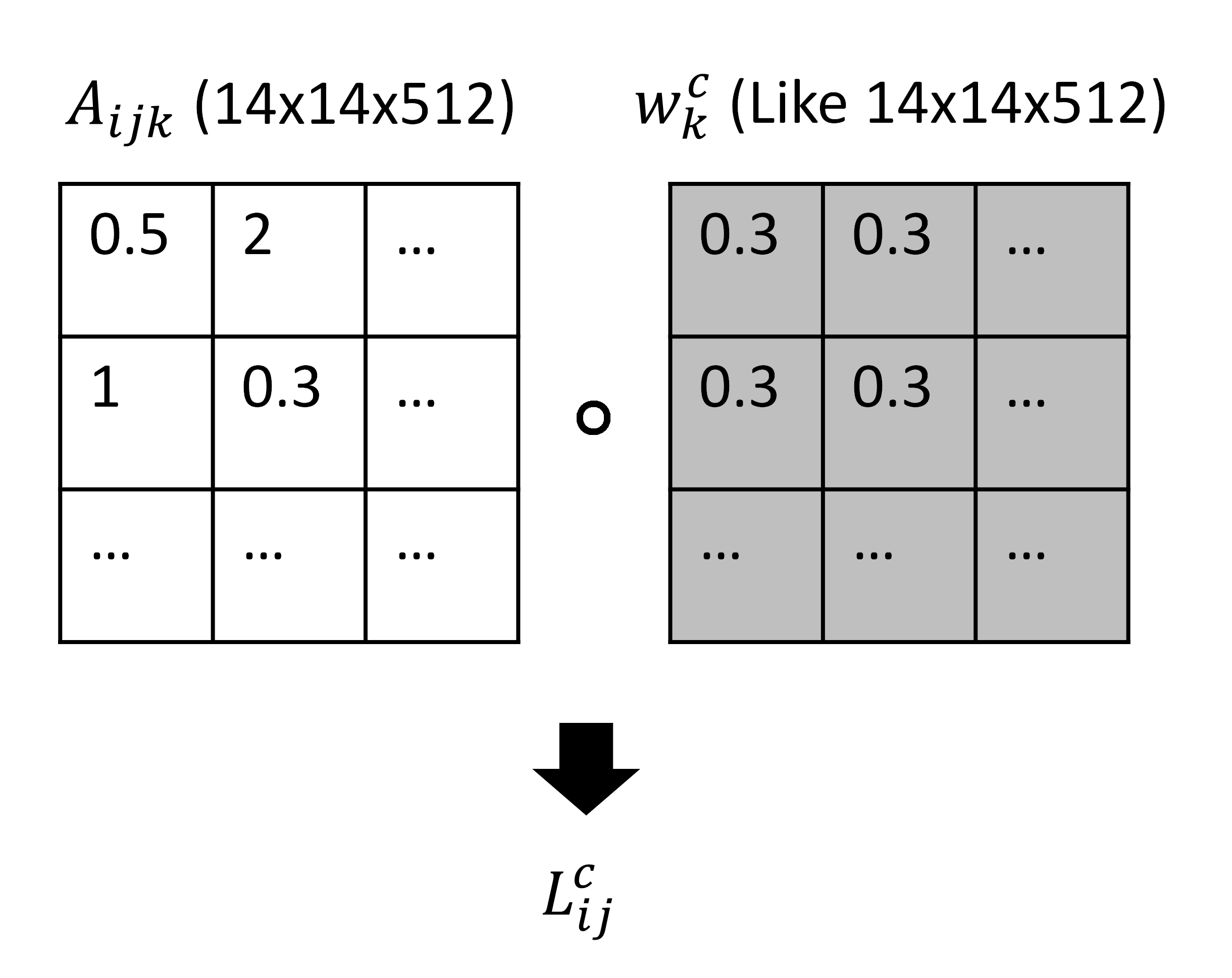} 
	\end{subfigure}
	\caption{Intuitive comparison of $w$ setting. (Left) Our Extended-CAM use $w_{ijk}^c$ that preserves $i, j$ information. Implicit smoothing does not occur. (Right) Existing CAM methods use averaged $w_k^c$. Implicit smoothing occurs in $L_{ij}^c$. The symbol $\circ$ means element-wise product, i.e., the Hadamard product.}
	\label{fig:intuitivecomparison}
\end{figure*}

Meanwhile, the visualizations of Extended-CAM tend to be wider than Grad-CAM and Grad-CAM++ (Fig. \ref{fig:visexamples}). The area of the CAM mask can be an important factor. Depending on the distribution of $L_{xy}^c$, the original image $I$ may be masked much or less masked. Thus, indicators such as Average Drop \% or \% Increase may be influenced. In the visualization task, the key is to find relatively important parts in the image, regardless of the absolute area of the mask. In consideration of this, we designed additional experiments with relative masking.

We designed the following experiments with relative masking, which masks with a fixed area. Based on the value of $L_{xy}^c$, we left only the top 50 \% pixels in the image and mask out the remaining 50 \% of pixels. Using masked images from relative masking, Average Drop \% and \% Increase are measured again (Table \ref{tab:compareresultsrelative}). Extended-CAM also exhibits better results than existing CAM methods in relative masking. This means that Extended-CAM finds important parts in the image well regardless of the distribution of $L_{xy}^c$.

\section{Discussion} \label{5 Discussion}

We proposed two options to improve the CAM method: 1) bilinear upsampling or Gaussian upsampling in the upsampling process, and 2) the existing formula using $w_k^c$ or the generalized one using $w_{ijk}^c$ in the $L_{ij}^c$ calculation. Apparently, the two options seem to be independent. However, in this section, we will discuss in detail that the two options are not independent and determine a factor, smoothness.

First, Gaussian upsampling has the effect of making $L_{ij}^c$ smoother. In terms of signal processing, Equation \ref{eqn:gaussianup} is equivalent to applying zero-insertion and filtering. The latter is commonly known as Gaussian smoothing. In another aspect, as mentioned in Section \ref{3 Method}, Gaussian upsampling differs from conventional upsampling methods and more naturally represents grid-pixel relationships. If we use Gaussian upsampling, a pixel-level CAM value is also influenced by distant grid cells. That is, Gaussian upsampling combines $L_{ij}^c$ through a wide area. Thus, when bilinear upsampling is replaced with Gaussian upsampling, $L_{xy}^c$ becomes smooth overall. We call this additional smoothing {\it explicit smoothing}.

Second, consider the setting of $w$. We illustrate an intuitive comparison of the use of $w_{k}^c$ and $w_{ijk}^c$ (Fig. \ref{fig:intuitivecomparison}). $w_{k}^c$ averages the gradient in the $i, j$ direction and ignores the $i, j$ units. Because $L_{ij}^c$ is calculated from $w_{k}^c$ and $A_{ijk}$, the averaged $w_{k}^c$ creates a smoothing effect on $L_{ij}^c$. We call this effect {\it implicit smoothing}. In contrast, if $w_{ijk}^c$ is used, implicit smoothing does not appear because $L_{ij}^c$ is calculated using the gradient that is not averaged in the $i, j$ direction.

In summary, the existing CAM method of using bilinear upsampling and $w_k^c$ causes implicit smoothing. Our method of using Gaussian upsampling and $w_{ijk}^c$ leads to explicit smoothing. Thus, both the upsampling method and the $w$ setting option determine smoothness. Indeed, Gaussian upsampling can mimic the effects of the use of $w_k^c$. We found that if we apply explicit smoothing using certain smoothness $\hat\sigma_X = \hat\sigma_Y = 20$ rather than the standard deviations of the effective receptive field, the results become quite similar to those of Grad-CAM where implicit smoothing occurs.

\begin{table*}[ht!]
	\centering
	\begin{tabular}{lrrrr}  
		\toprule
		Equation & Upsampling Method & \begin{tabular}{@{}c@{}}Average Drop \% \\ (Lower Better)\end{tabular} & \begin{tabular}{@{}c@{}}\% Increase \\ (Higher Better)\end{tabular} & Smoothing \\
		\midrule
		Grad-CAM & Gaussian upsampling & 43.64 & 17.46 & Implicit + Explicit \\
		         & Bilinear upsampling & 47.94 & 16.39 & Implicit \\
		Grad-CAM++ & Gaussian upsampling & 47.79 & 13.99 & Implicit + Explicit \\
		           & Bilinear upsampling & 45.60 & 15.10 & Implicit \\
		Extended-CAM & Gaussian upsampling & {\bf 39.86} & {\bf 20.72} & Explicit \\
		             & Bilinear upsampling & 79.23 & 4.02 & No \\
		\bottomrule
	\end{tabular}
    \caption{We examined various smoothing options. The best result can be obtained using explicit smoothing only. If we use other smoothing options, the results become worse due to the wrong smoothness. We measured Average Drop \% and \% Increase using relative masking.}
	\label{tab:combine}
\end{table*}

So, how should smoothing be applied? Note that implicit smoothing has a smoothing effect on $L_{ij}^c$ through averaging $w_k^c$. Here, smoothness is arbitrarily determined. If we use implicit smoothing, we won't be able to control the smoothness as desired, even if we know the reasonable smoothness. Explicit smoothing, in contrast, allows us to set the smoothness directly. Note that Gaussian upsampling naturally links the grid-pixel relationship and it means smoothing. This implies that the appropriate amount of smoothness exists and is determined by explicit smoothing. Therefore, it is preferable to apply explicit smoothing by calculating the appropriate smoothness rather than arbitrarily determining the smoothness through implicit smoothing. 

Indeed, the amount of smoothing strongly relates to the accuracy of the visualization mask. We believe that the reason Extended-CAM provides a more accurate visualization mask lies in the valid smoothness that explicit smoothing determines. We present the observations that other incorrect smoothnesses degrade performance (Table \ref{tab:combine}. We found that applying Gaussian upsampling to Grad-CAM/Grad-CAM++ results in worse visualization than Extended-CAM. This means that using both implicit smoothing and explicit smoothing at the same time applies the smoothing twice and determines the wrong smoothness. Also, when bilinear upsampling and $w_{ijk}^c$ are used instead of Gaussian upsampling and $w_k^c$, the performance is degraded. We believe these options also determine wrong smoothness because both implicit and explicit smoothings do not appear. These observations agree with our analysis that applying explicit smoothing alone determines valid smoothness, leading to improved performance.

\section{Conclusion and Future Work} \label{6 Conclusion}
This paper proposes Extended-CAM, a new CAM-based visualization method of DCNN. First, the limitations of bilinear upsampling are identified in the pixel-level upsampling. We discussed the validity of Gaussian upsampling using an effective receptive field. Second, the problems of the mathematical derivation presented in the previous papers are identified and corrected. We found that these two options are factors that determine smoothness and can be interpreted as explicit smoothing and implicit smoothing, respectively. Experimental results showed that Extended-CAM represents a more accurate visualization than previous studies.

There are still tasks left. All CAM approaches have a limitation in that the last feature map is highly coarse ($14 \times 14$). In this study, we focused on the last feature map. But in the future, we expect the finer structure in the image to be investigated by applying a CAM that sets another feature map in the front as a target layer. Also, while we used DCNN for 2D images, we expect the potentials of Extended-CAM for other general-purpose DCNNs such as speech recognition.

\bibliographystyle{named}
\bibliography{ijcai20}

\end{document}